\pdfoutput=1

\documentclass[11pt]{article}

\usepackage[final]{EMNLP2023}

\usepackage{times}
\usepackage{latexsym}

\usepackage[T1]{fontenc}

\usepackage[utf8]{inputenc}

\usepackage{microtype}

\usepackage{inconsolata}
\usepackage{graphicx}
\usepackage{float}
\usepackage{enumitem}
\usepackage{subcaption}
\usepackage{soul}

\newcommand{\sra}[1]{\textcolor{cyan}{#1}}
\newcommand{\sta}[1]{\st{#1}} 

\renewcommand{\sra}[1]{#1} 
\renewcommand{\sta}[1]{\!} 
\usepackage[T1]{fontenc}
\usepackage[utf8]{inputenc}
\usepackage{tikz}
\usetikzlibrary{shadows}
\newcommand*\keystroke[1]{%
  \tikz[baseline=(key.base)]
    \node[%
      draw,
      fill=white,
      drop shadow={shadow xshift=0.25ex,shadow yshift=-0.25ex,fill=black,opacity=0.75},
      rectangle,
      rounded corners=2pt,
      inner sep=1pt,
      line width=0.5pt,
      font=\scriptsize\sffamily
    ](key) {#1\strut}
  ;
}

\title{\includegraphics[height=\fontcharht\font`\B]{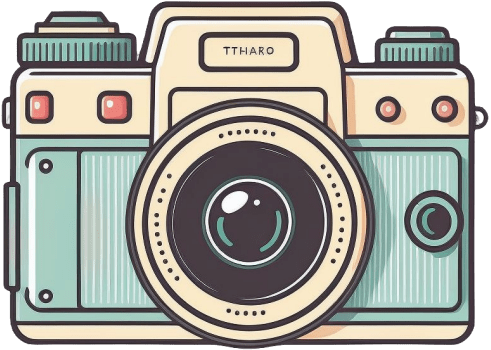} CAMRA: Copilot for AMR Annotation}

\author{Jon Z. Cai \and Shafiuddin Rehan Ahmed \and Julia Bonn \AND Kristin Wright-Bettner\and Martha Palmer \and James H. Martin\\
        University of Colorado Boulder \\ \texttt{jon.z.cai@colorado.edu}}

\begin{document}
\maketitle
\begin{abstract}
In this paper, we introduce CAMRA (\textbf{C}opilot for \textbf{AMR} \textbf{A}nnotatations), a cutting-edge web-based tool designed for constructing Abstract Meaning Representation (AMR) from natural language text. CAMRA offers a novel approach to deep lexical semantics annotation such as AMR, treating AMR annotation akin to coding in programming languages. Leveraging the familiarity of programming paradigms, CAMRA encompasses all essential features of existing AMR editors, including example lookup, while going a step further by integrating Propbank roleset lookup as an autocomplete feature within the tool. Notably, CAMRA incorporates AMR parser models as coding co-pilots, greatly enhancing the efficiency and accuracy of AMR annotators. To demonstrate the tool's capabilities, we provide a live demo accessible at: \url{https://camra.colorado.edu}\footnote{publish upon acceptance, demo video link: \url{https://youtu.be/mS3tzDVVaU8}}
\end{abstract}

\section{Introduction}
Abstract Meaning Representation (AMR) stands as one of the most widely embraced formalisms for deep lexical semantic representation within the NLP community. It effectively captures the lexical semantics present in multiple sentences by employing a directed, acyclic graph, wherein graph nodes form predicate-argument structures locally in Neo-Davidsonian fashion  \cite{banarescu2013abstract}. AMR can address both superficial semantic inquiries, encompassing aspects like "Who did what to whom, when, where, and how," as well as the intricate relationships between various events and states. Beyond these merits, AMR offers an invaluable advantage through its transparent symbolic representation of the semantics inherent in natural language text, significantly benefiting tasks reliant on semantic inference and necessitating interpretability.

Over the past decade, NLP researchers have meticulously transcribed tens of thousands of natural sentences into AMR graphs  \cite{LDC2014T12, LDC2017T10, LDC2020T02, may-2016-semeval, bonial-etal-2020-dialogue,bonn-etal-2020-spatial}, providing a critical source for the statistical machine learning approach to semantic parsing. 
While these resources are invaluable, producing AMRs is difficult, involving many sub-tasks, such as the annotation of nouns/named-entities, predicate-argument dependencies, co-reference resolution, discourse connectives, negation, and temporal relations. On top of this, annotators would need to be thoroughly trained to navigate a complex annotation tool interface in the process. 

Traditional AMR editors typically begin by having the annotator construct a root node and then add additional nodes as children through graph traversal. Nodes are added either through a dashboard made up of a combination of buttons, menus, and entry fields or, through a command-line-like interface with a series of learned commands. Both versions add yet another layer of learning complexity to the already intricate AMR structure. However, this is not simply an AMR problem but a problem for all semantic annotation tasks that involve complex structural layers of annotation.
\begin{figure}[H]
    \centering
    \includegraphics[width=\columnwidth]{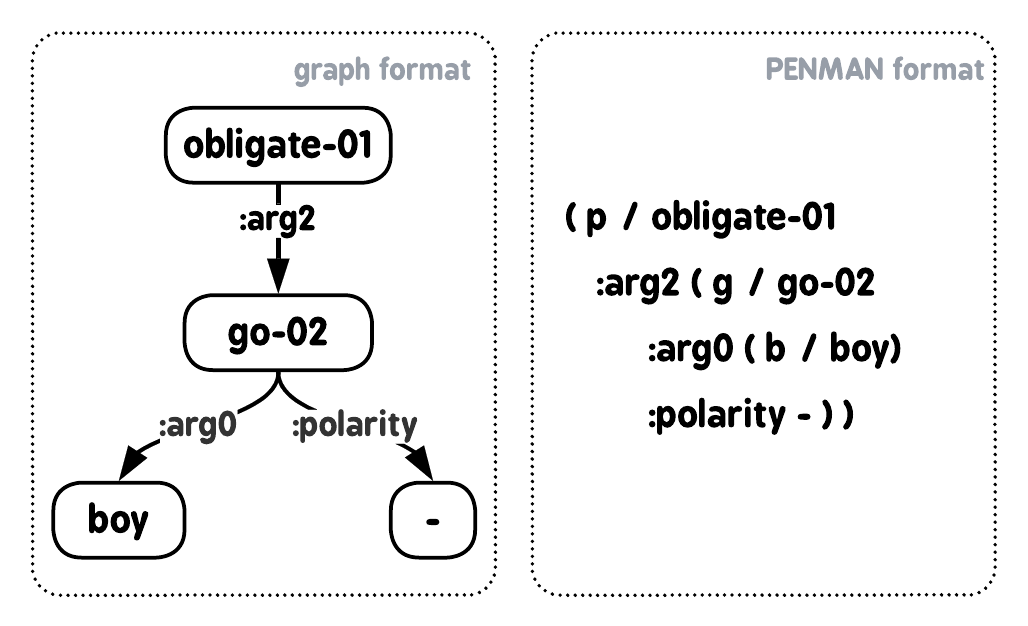}
    \caption{AMR for sentence "The boy must not go." in conventional graph representation format (left) and PENMAN encoding language format (right)}
    \label{fig:amr-example}
\end{figure}
We present an example AMR in Figure \ref{fig:amr-example} for the sentence ``The boy must not go.'' In an AMR graph, predicates and their corresponding arguments are represented by nodes. In this example, the \texttt{go-02} predicate\footnote{From PropBank, \url{https://propbank.github.io/}}  \cite{palmer2005proposition, pradhan-etal-2022-propbank} has one argument, which is \texttt{boy}. AMR specifies the role of each argument with labeled edges. Core roles, like stereotypical agent, patient, and thematic role, are typically denoted by \texttt{arg0}, \texttt{arg1}, and \texttt{arg2}, respectively. Other non-core roles, which are usually predicate-specific, are directly labeled with their names, such as \texttt{location}, \texttt{direction}, \texttt{time}, and \texttt{duration}. AMRs can be expressed in various formats, but graphically annotating their complex structure is impractical. To address this researchers adapted PENMAN notation  \cite{Goodman:2019, goodman-2020-penman}, which represents graph structures using bracketing syntax. Labeled edges are encoded with the preceding colon symbol, and opening brackets indicate new AMR nodes. Terminal nodes are denoted by closing parentheses. 

Advancements in large language model-based coding assistance, like Codex \cite{openai-codex} and Copilot by OpenAI and Microsoft have been revolutionizing program synthesis for software engineering tasks. These models are trained for both natural languages and programming languages, enabling them to intelligently complete programs based on code history and human instruction. Drawing inspiration from \sra{code-completion approaches}, we take a similar path by integrating an AMR parser model alongside a human annotator. This unique combination allows us to streamline the AMR construction process, handling easier yet tedious tasks like named entity sub-graph construction through the parsing assistant. At the same time, more intricate annotation decisions, such as predicate sense distinction, discourse connections, and co-reference resolutions, can be moderated by the annotators themselves, ensuring a balanced and effective approach to AMR annotation. \\
We summarize our contributions to the semantic annotation task as follows:
\begin{itemize}[leftmargin=*]
    \item We designed and implemented an innovative online AMR annotation tool that treats semantic annotation as a coding task streamlining the annotation process.
    \item We present the annotator-centric tool equipped with local Propbank snippet autocomplete and full generative model-based suggestions, enhancing the annotation experience for both beginner and experienced annotators.
    \item We introduce an intuitive click-based matching process for AMR concept alignment, simplifying and accelerating the alignment step for a smoother annotation experience.
\end{itemize}
Our highly modularized implementation enables easy swapping of language syntax and assistant models, creating a flexible "programming as annotation" paradigm adaptable to various languages and structures.
\section{Related Work}
The two most widely used AMR annotation tools are ISI AMR Editor (ISI-Editor) \cite{isi-editor} and Anafora \cite{chen-styler-2013-anafora}. Both are web-based text annotation tools that focus on different levels of AMR annotation. ISI-Editor is primarily designed for lexical-level AMR annotation, while Anafora is commonly used to construct document-level AMRs based on existing sentence-level AMRs. Our work with CAMRA is primarily comparable to ISI-Editor, as we also focus on sentence-level AMR construction. However, it is worth noting that editing cross-sentence relations, such as inter-sentential coreference resolution, can also be accomplished through CAMRA with relative ease. We will later showcase how to use ISI-Editor in comparison to our approach.\\
\begin{figure}[ht]
    \centering
    \includegraphics[width=\columnwidth]{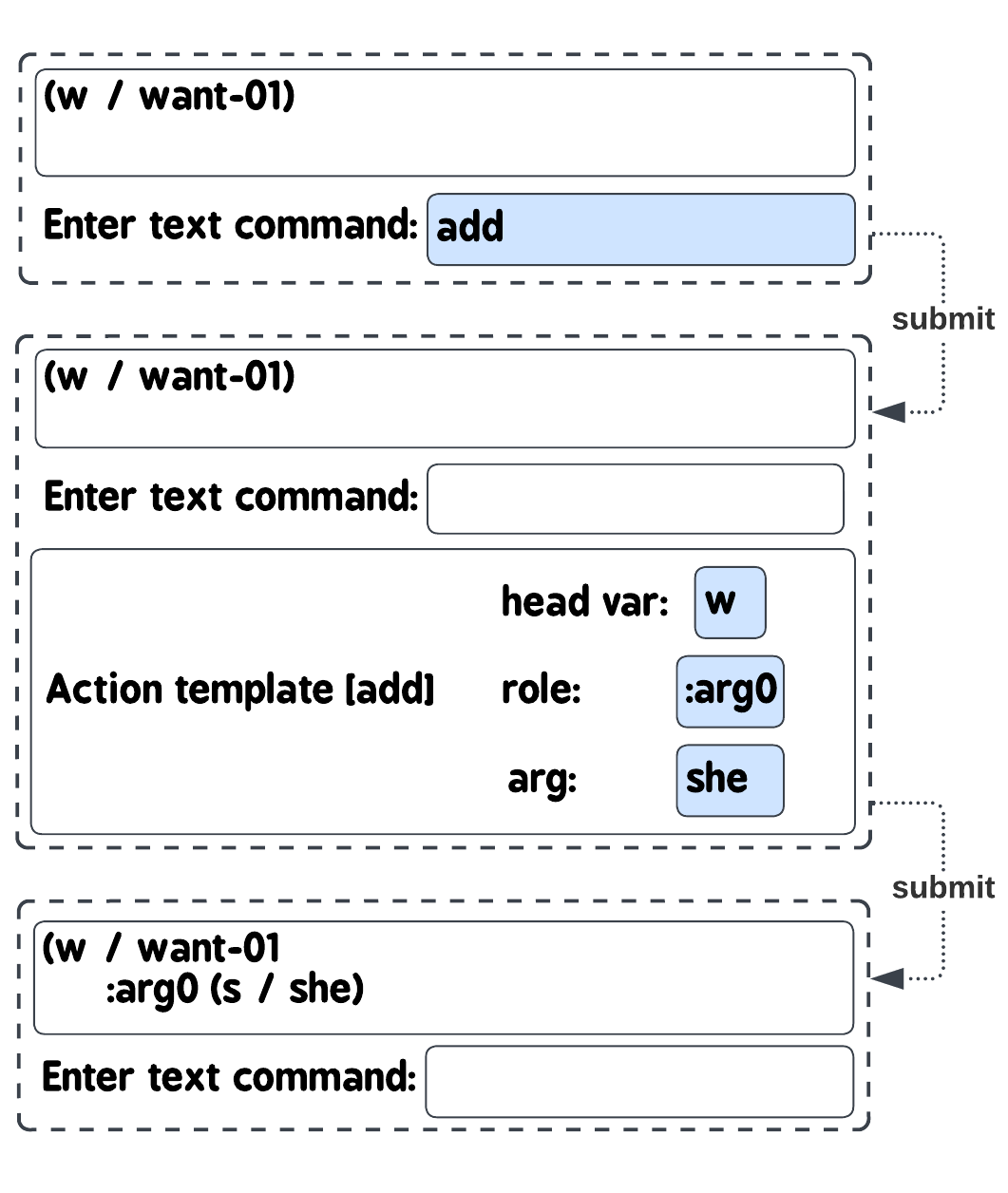}
    \caption{Adding a new argument to predicate node \texttt{(w / want-01)} with ISI editor's interface. Each dashed line box represents an updated view after submitting the mini form. Blue colored fields of each form represent fields that are required to be filled before submitting}
    \label{fig:isi-editor-procedure}
\end{figure}
The ISI editor offers comprehensive support for editing an AMR graph through various operators, including \textit{top} to initiate an AMR graph and \textit{add} to create an AMR triplet relation. Figure \ref{fig:isi-editor-procedure} illustrates an example of this functionality\footnote{To demonstrate the interaction of the interface in a straightforward manner, we use diagrams in Figure \ref{fig:isi-editor-procedure} instead of real screenshots. These diagrams faithfully represent the relative positions and interactive logic of the ISI editor.}. In the first view, we enter the \texttt{add} operator in the command field and submit it to activate the Action template view. Here, we proceed to fill in the new role with the specified head variable, role label, and argument concept node. ISI Editor processes the template form with verification to ensure the action is valid, resulting in an updated AMR displayed in the viewport. In total, annotators have access to 8 core operators that allow them to modify the AMR graph in PENMAN encoding form with shortcuts for advanced users. Additionally, the ISI Editor provides helpful annotation facilitations, such as the ability to search for existing AMR data and perform error checks on demand. 
\sra{These functionalities are organized within a dashboard interface, equipped with menu buttons that trigger specific features.}
Using ISI Editor becomes a process of sequentially filling small forms.

There are other annotation tools available for various complex linguistic-driven tasks, such as the UCCAApp \cite{birch-etal-2016-hume} for Universal Conceptual Cognitive Annotation \cite{abend-rappoport-2013-universal}, the brat rapid annotation tool \cite{brat-at} for universal dependency tree construction, TreeEditor for Rhetorical Structure Theory \cite{pajas-stepanek-2008-recent}. However, unlike AMR, which can exist without explicit alignment of concept nodes to the natural language surface text, these mentioned tasks are tightly anchored to the surface text. Consequently, they require click-selection-based interactions with the user to initiate the annotation process. Knowtator is another annotation tool that facilitates ontology construction in Protege from text, the UI design of Knowtator \cite{ogren-2006-knowtator} also relies on small form filling.   

Furthermore, most syntactic tasks involve a limited number of relationships, typically not exceeding a dozen, in contrast to AMR, where the number of rolesets directly corresponds to the number of predicates in a given language. In languages like English, the number of predicates can easily surpass 5000. The unique challenge of annotating AMR, coupled with the lack of support for other formalisms, has motivated us to create an annotation tool equipped with a formal language coding environment. This tool aims to enhance the efficiency and accuracy of AMR annotation and provide a novel solution to handle its distinctive complexities.

Recently, significant strides have been made in advancing the development of the model-in-the-loop annotation style, aimed at fostering machine-assisted human annotations. Popular tools, such as Prodigy\footnote{www.prodi.gy} and INCEpTION  \cite{tubiblio106270}, primarily focus on providing annotation suggestions for Text Classification, Named Entity Recognition, and Entity Relation Extraction. More recent methodologies have expanded these capabilities to encompass entity and event coreference resolution  \cite{bornstein-etal-2020-corefi, ahmed-etal-2023-good}. However, the field of machine-assisted annotations for AMR remains relatively under-explored. Our work endeavors to address this gap, contributing to the enhancement and expansion of this vital aspect of the annotation landscape.
\section{Design and Features}
\begin{figure*}[t]
    \centering
    \includegraphics[width=\textwidth]{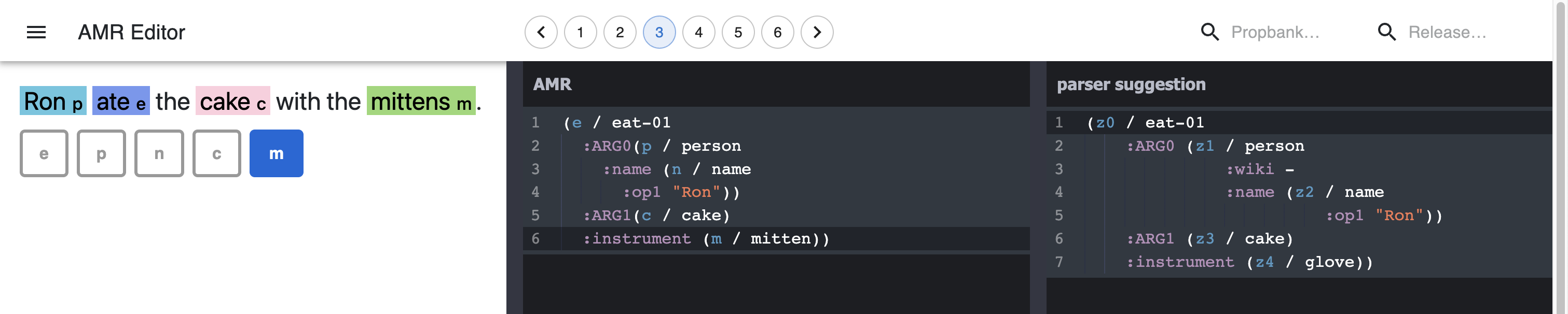}
    \caption{an overview of the CAMRA editor with an annotated example sentence. Left panel is the surface text area with dynamic variable carryover from the constructed AMR code. The middle panel is the main AMR editing area where the string complies with the PENMAN encoding syntax for AMR. The right text panel renders the parser suggestions. Note: this screenshot contains only nonempty part of the UI, the UI is window size responsible. }
    \label{fig:CAMRA-overall}
\end{figure*}
To enhance the effectiveness of annotator-computer interaction, it is essential for the computer to play an active role in the annotation process, rather than serving merely as a passive typewriter. At the same time, it is important to minimize the need for annotators to frequently shift their attention among different views to memorize local predicate-argument structures temporarily while completing AMR annotations. To address these concerns, we have formulated the following design principles:
\begin{itemize}[leftmargin=*]
    \item Upon looking up dictionaries like Propbank rolesets or existing annotation examples, the annotator can take advantage of two options. Firstly, they can directly invoke the desired frame within the coding environment, leading to the automatic completion of the target structure. Alternatively, they can easily copy relevant sections from examples and paste them into the coding environment, streamlining the annotation process.
    \item The produced AMR should undergo active parsing to ensure its legality and provide valuable feedback to the annotators.
    \item To make the annotation experience akin to coding, the editor needs to incorporate additional text editing tool features, such as multiple selections, code difference highlight, and editing capabilities, thereby optimizing the annotators' workflow and overall experience.
    \item The design of the copilot editing environment should be versatile and adaptable to different annotation projects. It should support a general-purpose approach, enabling similar annotation tasks to be accomplished with ease by merely switching formal language syntax as needed.
\end{itemize}

\subsection{Features}
We show the main app interface in Figure \ref{fig:CAMRA-overall}.
\subsubsection{Annotation panels}
CAMRA is primarily composed of three horizontally laid-out panels. The leftmost panel is designed for rendering the surface text and aligning AMR concept nodes to the corresponding text. This panel displays two blocks of information: the surface text itself and the AMR node variable names present in the middle panel.

The middle panel serves as the AMR text editor, equipped with common code editor features, such as syntax highlighting, auto bracket matching and closing, and snippet auto-complete. Writing AMR in this text editor closely resembles writing code in a programming language.

Finally, we utilize the right panel to render the AMR generated by the parser. Annotators have the option to use any part of the parser-suggested AMR by simply copying the text over to the middle panel, facilitating a seamless integration of the parser's suggestions into the annotator's workflow. This three-panel layout ensures a smooth and intuitive annotation process for CAMRA users.

\subsection{Autocomplete}\label{sec:autocomplete}
CAMRA is equipped with two levels of auto-completion mechanisms: local autocomplete and global autocomplete. The local auto-complete feature considers only the nearest string to provide suggestions for reserved keywords and Propbank templates. This proves useful in cases where AMR relation prompting is required, as it relieves annotators from the burden of remembering every relation precisely. This is particularly helpful for non-core AMR relations, which can be quite lengthy and prone to errors. Additionally, local autocomplete is computationally less intensive compared to global autocomplete, utilizing substring matching as the search algorithm and edit distance as the ranking algorithm when invoked.

Moreover, when snippet autocomplete is activated, the editor holds field-like text spans in memory, allowing annotators to simply type in the value and switch to the next field by pressing the \keystroke{~Tab~} key, significantly reducing navigation time within the code.

In contrast, the global autocomplete from the machine learning-based parser considers both the surface text and its generated history, making it more comprehensive than the local autocomplete. However, due to its higher computational cost, the parser suggestion is invoked only once per sentence. We keep the parser suggestion API open to backend updates, enabling the possibility of further tailored parser suggestions that take the users' input into account for even more personalized and refined suggestions.

\subsection{Manual Search}
\begin{figure}[t!]
    \centering
    \includegraphics[width=1.05\columnwidth]{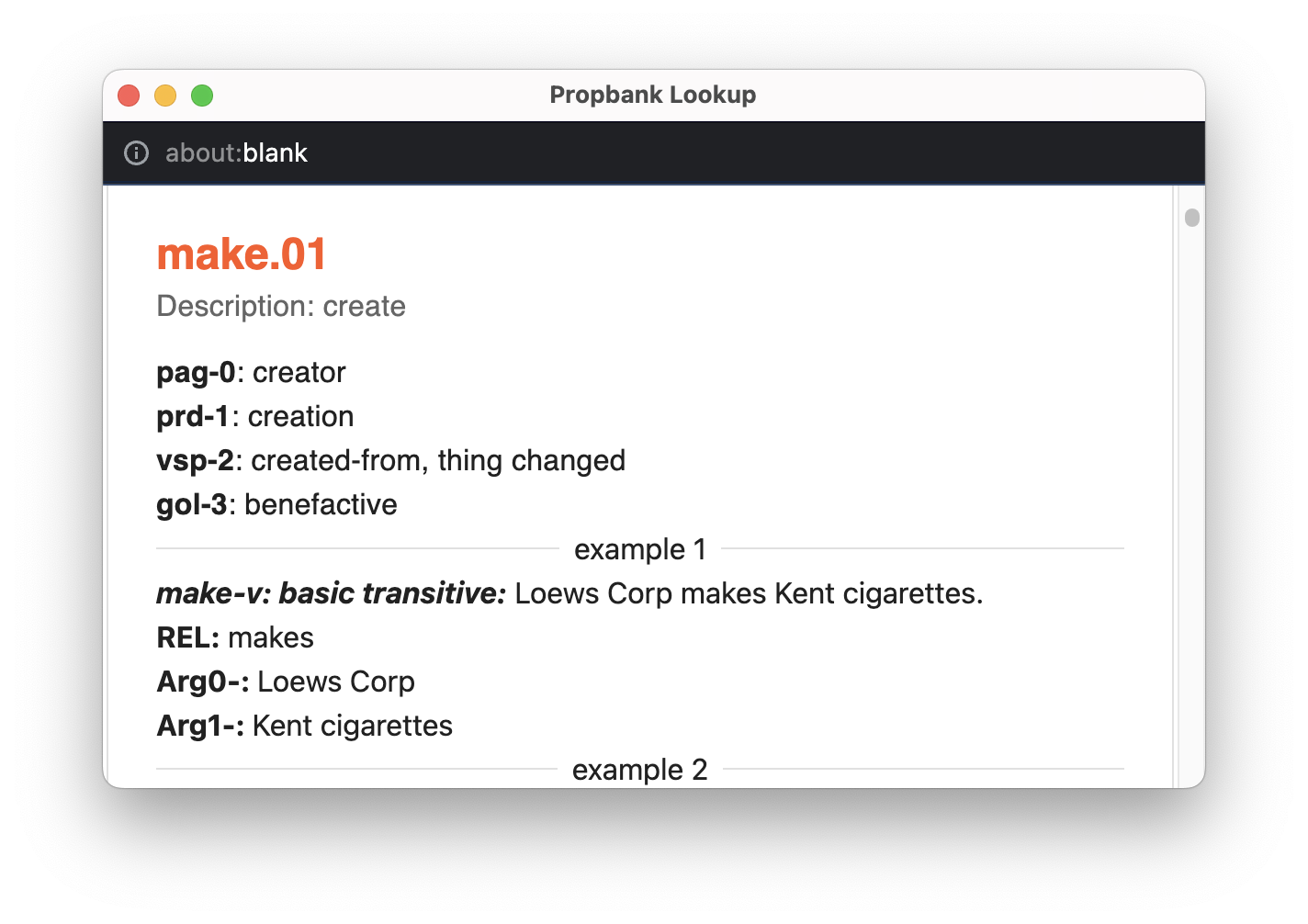}
    \caption{When looking up in the Propbank rolesets for the keyword "make," a persistent new window will appear at the annotator's disposal on the side of the CAMRA's main interface.}
    \label{fig:propbank-search}
\end{figure}

\begin{figure}[ht]
    \centering
    \includegraphics[width=\columnwidth]{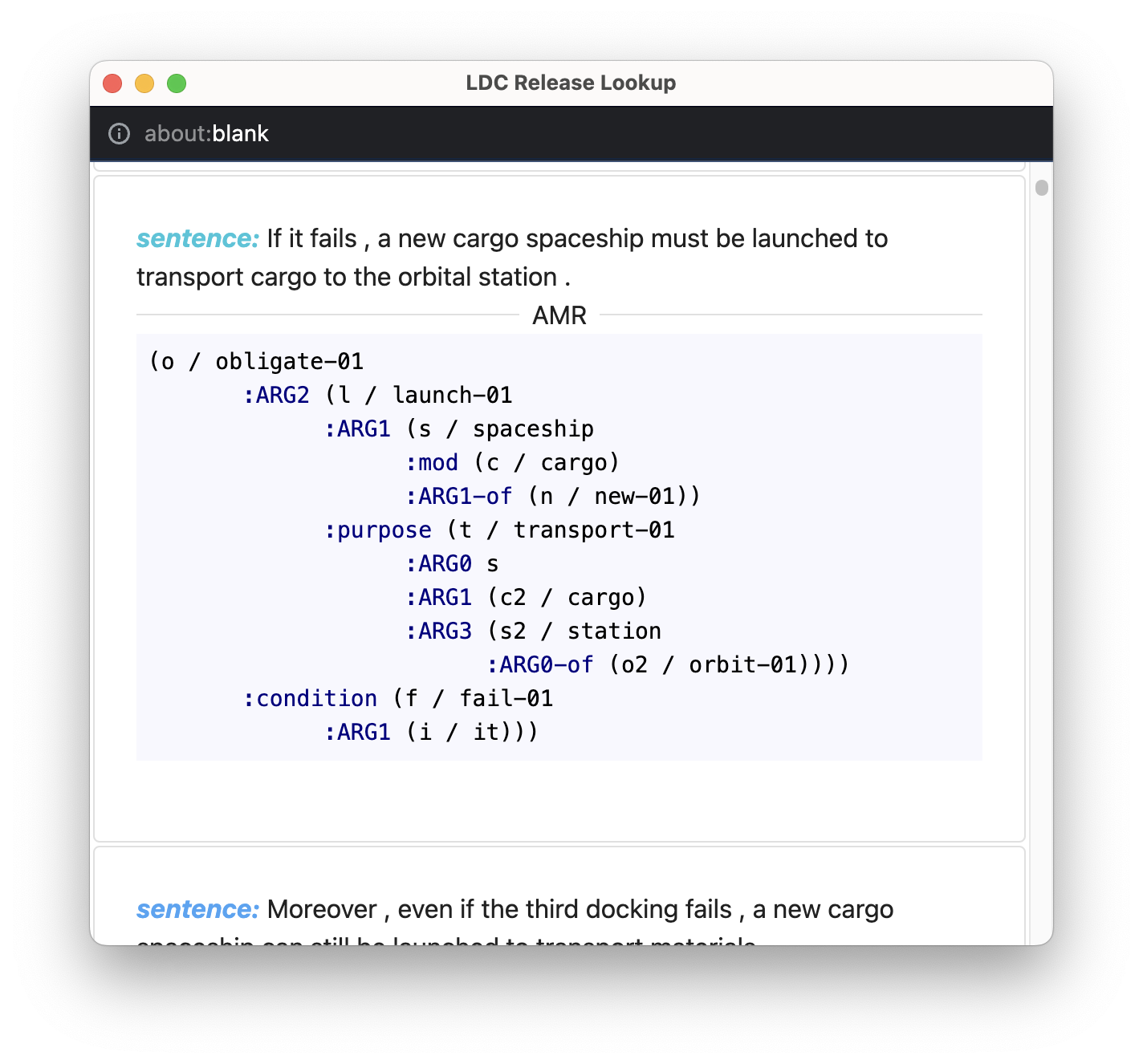}
    \caption{When looking up in the existing AMRs corpora for the keyword "must", another persistent new window will appear at the annotator's disposal on the side of the CAMRA's main interface. }
    \label{fig:release-search}
\end{figure}

CAMRA also incorporates a similar search function for both Propbank and the existing AMR corpus. This feature functions similarly to the ISI editor's search, popping up with a more updated UI design when invoked. In Figure \ref{fig:propbank-search} and Figure \ref{fig:release-search}, we demonstrate the search results for the keywords ``make'' in Propbank and "must" in LDC2020T02 AMR corpus \cite{LDC2020T02}.

To ensure a clutter-free main annotation window, we have dedicated individual windows to host the search results. These search windows persistently update their content whenever a new search is launched. Annotators can easily browse and perform \keystroke{ Ctrl+f } searches within these windows.

In organizing and highlighting the AMRs in the existing corpus search, we have maintained the same format as the main AMR editing panel. This facilitates straightforward copy-pasting of any desired part of the AMR into the AMR editing panel, providing annotators with direct access to the information they need for a more efficient and streamlined annotation process.

\begin{figure*}[h]
    \centering
    \includegraphics[width=\textwidth]{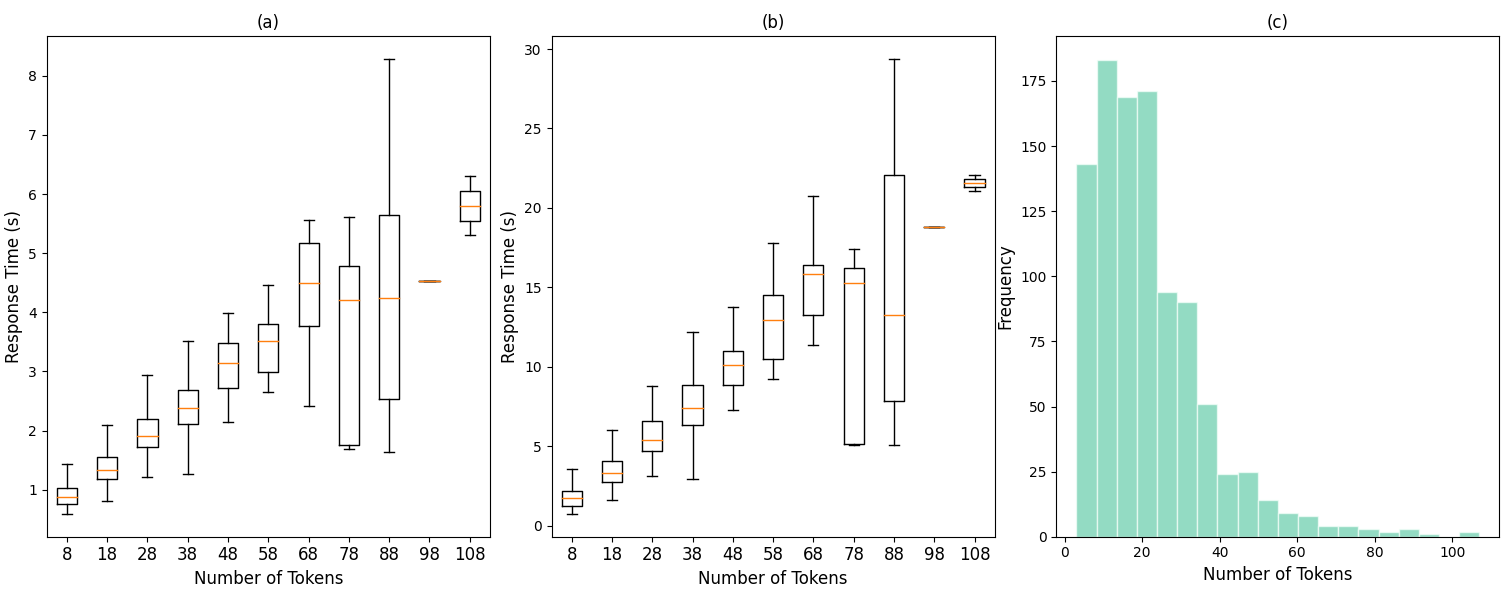}
    \caption{An overview of the AMR parser server's response time is shown for the same 1000 randomly selected sentences from the LDC2020T02 AMR corpus training set. Figure (a) displays the response time box chart without GPU support, while Figure (b) shows the response time box chart with a single GPU support. Additionally, Figure (c) represents the frequency distribution of sentence lengths among the 1000 sentences. The red middle line of each box candle represents the median, the box specifies the interquartile range (IQR), and the whiskers indicate the 1.5 IQR range.}
    \label{fig:sample-response-time}    
\end{figure*}
\subsection{Utility Functions}
All managerial and administrative functions are conveniently placed in a hidden menu accessible through the top left corner drawer icon. This menu houses various actions, including uploading a new workset (a text file containing all the target text to be annotated), uploading an annotation checkpoint, profile management, and more. As these functionalities are not the primary focus of our CAMRA and do not represent critical components for this paper, we have opted to exclude them from further discussion but let the reader explore in demonstration.
\subsection{Language Servers}
The core active assistance feature of our annotation tool is powered by language servers on the backend. As elaborated in Section \ref{sec:autocomplete}, we have two layers of language support: a local one and a global one, achieved through two REST-API servers.

Handling managerial tasks such as login, data storage, Propbank and release searching, and parser inquiries is a Django \cite{django} REST API server's responsibility. In addition, we have set up a separate REST API server dedicated to hosting pre-trained AMR parser models. This division allows for enhanced flexibility in resource distribution. For example, the managerial server can efficiently manage data transactions from the front end without requiring GPU support. On the other hand, most state-of-the-art AMR parser models are large neural network models that greatly benefit from GPU or other accelerating devices' computational power.

The design of CAMRA revolves around modularity as a critical principle, enabling the easy integration of various assistant models. For instance, when annotating unique domains of text, parsers previously trained on different domains may perform poorly, limiting the support they can offer to annotators. By being modular, our system can readily adapt to such scenarios. Additionally, this modularity facilitates the distributed deployment of our system.

At present, we offer support for the SPRING AMR parser \cite{bevilacqua-etal-2021-one} trained on LDC and spatial AMR corpus \cite{bonn-etal-2020-spatial} respectively as parsing assistance. However, the flexibility of our design makes it possible to include other assistant models tailored to specific needs in the future. The base model of the SPRING parser is BART-large \cite{lewis-etal-2020-bart} with nearly 140M parameters and requires approximately 2.2 GB of memory for inference. The integration of the parser facilitates model-in-the-loop learning processes, which can be adapted based on user requirements without significant difficulty.

\section{Discussion}
The primary user experience factor for our annotation tool is the response time of the parser-based language server. To evaluate this response time in relation to sentence length, we conducted tests using 1000 randomly selected sentences from the training set of LDC2020T02. The results are depicted in Figure \ref{fig:sample-response-time}. The average sentence length among these sentences is 21.72 tokens (tokenized by BART tokenizer), with a standard deviation of 14.76. With and without GPU support, the average response times are 1.62 and 4.47 seconds, respectively, with corresponding standard deviations of 0.86 and 3.32 seconds. We present a box chart illustrating the response time distribution in relation to sentence length (token numbers $n$), considering the BART model's theoretical inference complexity of $\mathcal{O}(n^2)$ and the prior distribution of sentence lengths. The testing machine has 2.2GHz Intel Xeon (R) CPU (24 cores), 256GB RAM and a NVIDIA Titan Xp GPU (12GB).
\section{Conclusion and Future Works}
CAMRA introduces a novel semantics annotation paradigm with considerable potential for enhancement. Given the similarity of the parser structure, integrating LLM into autocomplete and suggestion output is seamless. We are actively working on fine-tuning LLMs to make parsing copilot suggestions more interactive. To assess the language server's impact compared to traditional AMR annotation tools, we will conduct a human study involving annotators. Furthermore, we aim to expand the pool of potential annotators, serving the dual purpose of broadening our annotator base and supporting computational semantics education. Collaborating with NLP communities, we plan to extend support to other formalisms and annotation schemes. Moreover, we envision the integration of large language models into the language server, providing more natural language assistance from AI. This advancement could lead to yet another valuable application of large language models, enhancing their interpretability and error resilience through the fusion of neural and symbolic approaches. Such developments offer exciting possibilities for safer and more innovative applications.
\section*{Limitations}
CAMRA's language server support may encounter biases or challenges related to domain shift, as the underlying model's training data could be skewed towards specific text domains(such as newswire text). This might result in inaccuracies or reduced performance when dealing with text from different domains. Furthermore, while AMR serves as a versatile formalism, our PENMAN syntax design predominantly caters to English, potentially limiting its effectiveness for other languages. Expanding the PENMAN syntax to encompass a wider array of languages would not only improve its cross-linguistic applicability but also enhance the overall usability and inclusivity of the annotation tool.
\section*{Ethics Statement}
In addition to the limitations highlighted in the previous section, CAMRA has a core objective of enhancing human-computer communication through UI design and AI assistance. An essential aspect of this endeavor is to ensure that CAMRA users have a comprehensive grasp of how the language server operates and how it impacts annotations. This is achieved through transparent documentation and the provision of mechanisms for understanding the tool's decision-making process. Furthermore, we place significant emphasis on effective communication with annotators to consider the cultural and domain-specific sensitivities inherent in the text being annotated. Recognizing these nuances is crucial, as any misinterpretation or misrepresentation of cultural contexts could result in erroneous semantic annotations.
\section*{Acknowledgement}
This research was supported by the NSF National AI Institute for Student-AI Teaming (iSAT) under grant DRL 2019805. The opinions expressed are those of the authors and do not represent views of the NSF. The authors extend their heartfelt gratitude to Adam Zheng, Brad Johnson, Skatje Myers, Elizabeth Spaulding and Jie Cao for their unwavering support in handling hosting technicalities. Additionally, Dr. Alexis Palmer's suggestions in shaping the user study design. We thank the infrustructure and computational resources support from Chameleon Cloud \cite{keahey2020lessons} to make this project possible.

\end{document}